\documentclass[10pt, a4paper]{article}
\usepackage{lrec}
\usepackage{multibib}
\newcites{languageresource}{Language Resources}
\usepackage{graphicx}
\usepackage{tabularx}
\usepackage{soul}
\usepackage[usenames, dvipsnames]{color}
\usepackage{xcolor}
\usepackage{booktabs} 
\usepackage{amsmath}
\usepackage{epstopdf}
\usepackage[latin1]{inputenc}
\usepackage[noend]{algorithm2e}

\usepackage{hyperref}
\usepackage{xstring}
\usepackage{subcaption}
\usepackage{url}
\usepackage{color}
\usepackage{array, caption, tabularx, makecell, booktabs}%
\usepackage{bold-extra}

\title{SlugNERDS: A Named Entity Recognition Tool for\\ Open Domain Dialogue Systems}

\name{Kevin K. Bowden, Jiaqi Wu, Shereen Oraby, Amita Misra, and Marilyn Walker}

\address{Natural Language and dialogue Systems Lab\\
 University of California, Santa Cruz\\
         \{kkbowden, jwu64, soraby, amisra2, mawalker\}@ucsc.edu\\}

\newcommand{\ct}[2]{\textbf{\color{#1}#2}}

\abstract{In dialogue systems, the tasks of named entity recognition (NER) and named entity linking (NEL) are vital preprocessing steps for understanding user intent, especially in open domain interaction where we cannot rely on domain-specific inference. UCSC's effort as one of the funded teams in the 2017 Amazon Alexa Prize Contest has yielded Slugbot, an open domain social bot, aimed at casual conversation. We discovered  several challenges specifically associated with both NER and NEL when building Slugbot, such as that the NE labels are too coarse-grained or the entity types are not linked to a useful ontology. Moreover, we have discovered that traditional approaches do not perform well in our context:  even systems designed to operate on tweets or other social media data do not work well in dialogue systems. In this paper, we introduce Slugbot's Named Entity Recognition for dialogue Systems (SlugNERDS), a NER and NEL tool which is optimized to address these issues. We describe two new resources that we are building as part of this work: SlugEntityDB and SchemaActuator. We believe these resources will be useful for the research community.}

\begin{document}

\maketitleabstract

\section{Introduction}
\label{sec:intro}
 
\begin{table*}
\begin{tabular}
{@{}p{0.5cm}|p{8cm}|p{3in}@{}}
\toprule
1 & 
\textbf{User:} {Let's talk about comic books. Who is your favorite character?} & In open domain dialogue, the user can engage us on any topic. \\
\hline
2 & 
\textbf{System:} {I like [{\color{purple} Moon Knight}]$_{Thing}$ because [{\color{purple} he}]$_{Thing}$ can see the future sometimes.} & Fictional entities, such as obscure comic book characters, are classified as \textit{Thing}. Using our web based NEL, we could also summarized he is a \textit{Fictional Super Hero}.\\
\hline
3 & 
\textbf{User:} {I guess that is a cool power, when did [{\color{purple} he}]$_{Thing}$ first appear?} & SlugNERDS performs anaphoric resolution to perform discourse level entity linking, in this case associating \textit{he} with \textit{Moon Knight}. \\
\hline
4 & 
\textbf{System:} {I think [{\color{gray!90} Marvel Comics}]$_{Corporation}$ debuted [{\color{purple} Moon Knight}]$_{Thing}$ in [{\color{purple} werewolf by night}]$_{Thing}$ number 32} & In SlugNERDS, correctly identifying \textit{Moon Knight} allows us to perform web level entity linking - we have a link to a relevant Wikipedia article where we find more information about the character, specifically the series he first appeared in, \textit{Werewolf by Night}. \\
\hline
5 & 
\textbf{User:} you know I kind of like scary space \ct{blue}{movies}. I saw [{\color{blue} alien covenant}]$_{Movie}$ the other day & The \textit{movies} keyword matches directly to an entity type, thus increasing the systems probability of picking the correct movie entity.\\
\hline
6 & 
\textbf{System:} I thought [{\color{blue} Alien Covenant}]$_{Movie}$ was scary and action packed. & Often times users are most interested in discussing trending entities, \textit{Alien Covenant} had been released in the theaters recently.\\
\hline
7 & 
\textbf{User:} That's cool, have you \ct{blue}{seen} the new [{\color{blue} Star Wars}]$_{MovieSeries}$ \ct{blue}{movie}? & Both the \textit{movie} keyword and the \textit{seen} action are useful to detect all \textit{Movie} type entities, such as \textit{MovieSeries}, \textit{MovieClip}, and \textit{Movie}.\\
\hline
8 & 
\textbf{System:} I haven't had a chance yet. Which is your favorite in the series? & Differentiating between \textit{Movie} and \textit{MovieSeries} allows us to curate better follow up question.\\
\hline
9 & 
\textbf{User:} I think my favorite [{\color{blue} Star Wars}]$_{MovieSeries}$ \ct{blue}{movie} is [{\color{blue} revenge of the sith}]$_{Movie}$ & We just need the partial title, \textit{revenge of the sith}, to correctly identify the entity \textit{Star Wars: Episode III - Revenge of the Sith}. \\
\hline
10 & 
\textbf{System:} Who is your favorite actor in [\color{blue} {it}]$_{Movie}$? & We know that \textit{Movies} have \textit{actors}.\\
\hline
11 & 
\textbf{User:} I'm \ct{PineGreen}{watching} the [\textcolor{PineGreen}{Lakers}]$_{SportsTeam}$ play right now, which is funny because I hate [{\color{WildStrawberry} Los Angeles}]$_{City}$, even though I \ct{orange}{live} in [{\color{orange} California}]$_{State}$ - best \ct{orange}{state} in the [{\color{cyan} United States}]$_{Country}$. & We note the various levels of granularity, instead of just a location tag we can detect each destination with more precision.\\
\hline
12 & 
\textbf{System:} Whose your favorite player on the [\textcolor{PineGreen}{Lakers}]$_{SportsTeam}$?  & Using the Schema.org attributes, we know that \textit{SportsTeams} have \textit{athletes}.\\
\hline
13 & 
\textbf{User:} [{\color{YellowOrange} albert einstein}]$_{Person}$ is very smart [{\color{YellowOrange} he}]$_{Person}$ is my favorite scientist. I wonder if [{\color{YellowOrange} he}]$_{Person}$ would have liked \ct{brown}{reading} [{\color{brown} the hunger games}]$_{BookSeries}$ [{\color{brown} mockingjay}]$_{Book}$. & Note that users can abruptly shift to a different context at any time. In this example, the \textit{mockingjay} movies scored higher than the books initially, however once we take into account the expectation of Book type entities via the \textit{reading} action, the correct entities are selected.\\
\bottomrule        
 \end{tabular}
 \caption{Sample dialogue with entities highlighted in addition to SchemaActuators annotations.} \label{table:monsters}
\end{table*}

When building dialogue systems, reliable named entity recognition (NER) and named entity linking (NEL) are vital to understanding user intent, especially if these dialogue systems are open domain and intended to support conversations on any topic. In designing our open domain social bot, Slugbot\cite{Bowden17Slugbot}, for the 2017 Amazon Alexa Prize Contest\cite{AlexaPrize}, we discovered several challenges specifically associated with both NER and NEL. This paper discusses these challenges, and shows how we address them with Slugbot's Named Entity Recognition for Dialogue Systems (SlugNERDS), a tool designed for NER and NEL in open domain dialogue. Additionally we present two corpora, SlugEntityDB and SchemaActuator, which are based on over 10,000 real user conversations with the system. We perform an extensive analysis of our system and the corpora to identify important areas of future work.

NER and NEL have been actively researched topics for decades \cite{Finkel09StanfordNER,RatinovRo09RothINER,Ritter11TwitterNER,Derczynski15NerResources,GuptaSiRo17RothNEL}. However, the resulting entity classification is often coarse and does not encode an ontology.  For example, Stanford NER features only a small number of abstract entity types such as PERSON, LOCATION, ORGANIZATION, and MISC \cite{corenlp,Finkel09StanfordNER}; these categories don't provide enough information for dialogue interpretation and generalization. Although other resources such as that from Ratinov and Roth \shortcite{RatinovRo09RothINER} utilize additional external knowledge by extracting 30 gazetteers from both the web and Wikipedia, the entity types are still not as varied as we need, and the framework lacks a clear ontology. Furthermore, the alignment of classes between systems can be inconsistent as there is no universally shared  taxonomy between them and the various data streams necessary to support open domain conversation\cite{bowden2017combining}. While Ling and Weld \shortcite{Ling12FinegrainNER} attempt to address this by using 112 fine-grained entity types consistent with Freebase \cite{Freebase}, Freebase is no longer maintained and recent inspection has shown it to be significantly incomplete \cite{Dong14GoogleKnowledgeFusion}.

While the accuracy of these state of the art NER systems can be quite high, ranging between 80-90\% on long text, on short informal text, such as tweets, accuracies drop to between 30-50\% \cite{Derczynski15NerResources}. Tweets are much more representative of the data we see from users in an interaction with a social bot than newswire data. Specifically, utterances tend to be short, and due to the open domain setting, relevant context is not guaranteed.

However, unlike tweets, a dialogue system must maintain a discourse model which can sustain multi-turn dialogue for the duration of the user interaction. In fact, it is clear that a dialogue system, which necessarily must operate in real time, has inherent challenges not present in other settings. Real time systems must be optimized such that they function without significant response delays between turns; such delays can be introduced by  approaches that rely on running a  machine learning classifier in real time. Additionally, new named entities are very commonly discussed in a social setting and need to be recognized as trends and current popular topics change: this requires  systems to be constantly (nightly) retrained on newly annotated data. Moreover, there must be significant noise tolerance - in a social setting it is often the case that users will speak more informally; the spoken domain can be even more challenging as an entity resolution tool must also account for automatic speech recognition (ASR) misinterpretations. 

\section{Overview of Tools and Corpora}
\label{sec:overview}

To address these challenges we present Slugbot's Named Entity Recognition for dialogue Systems (SlugNERDS), an NER and NEL tool which leverages the Google Knowledge Graph API in conjunction with the Schema.org taxonomy to identify known entities. The tool is optimized with respect to noisy open domain conversation and is able to perform both discourse and web based entity linking. Table \ref{table:monsters} represents an annotated conversation based on real interactions with Slugbot. While one could enhance the result further by utilizing the state of the dialogue system to set system expectations, we are interested in evaluating our tool without making any assumptions of the system using it.  

To supplement SlugNERDS we also present SlugEntityDB, an annotated corpus which can be used to evaluate our system. This represents to our knowledge the first Schema.org entity type annotated corpus for this task. The SlugEntityDB contains 2100 samples, 500 taken directly from real user data collected by SlugBot\cite{Bowden17Slugbot} during the inaugural Amazon Alexa Prize contest, 1600 synthesized such that we can easily verify the richness of the corpus. Since we are operating in the open domain, it is also very likely that adding synthesized data will result in entities which have never been seen by the system previously. In this dataset the utterances are formatted similar to the input which a spoken dialogue system would receive from a state of the art ASR system. Each utterance is annotated in tuples which includes the direct strong overlap, entity types, and full entity name as per the Google Knowledge Graph. Table \ref{table:slugentity_example} includes a sample of this dataset.

\begin{table}[h]
\centering
\begin{tabular}{{@{}p{3.7cm}|p{3.7cm}@{}}}
\toprule
Utterance & Annotation \\ \hline
the lord of the rings was my childhood & (lord of the rings, MovieSeries Thing, The Lord of the Rings)\\ \hline
my favorite star wars movie is probably revenge of the sith & (revenge of the sith, Movie Thing, Star Wars: Episode III $\-$ Revenge of the Sith); (star wars, MovieSeries Thing, Star Wars)\\ \hline
i want to visit black mountain & (black mountain, TouristAttraction Mountain Place Thing, Black Mountain)\\
\bottomrule         
\end{tabular}
\caption{Samples from the SlugEntityDB.}
\label{table:slugentity_example}
\end{table}


Additionally, we provide the SchemaActuators corpus, a partially hand annotated probabilistic mapping between actions/specific keywords/phrases and entity classifications (such as indicating \textit{watch} or \textit{seen} are related to \textit{Movie} entities). An example of these mappings can be seen in Table \ref{table:monsters}, where appropriate entries in the corpus are colored to match the associated entity. Currently the verbs used in this corpus have been hand annotated. These seed verbs are then expanded using synonym relations from Wordnet\cite{wordnet} and other lexical resources. Using these verbs in addition to prepositional phrases we automatically generate a list of short phrases associated with specific entities, such as \textit{arrive at} for the \textit{City} entity. We use a similar process of automatic expansion on the entity type to generate a list of candidate keywords which can potentially indicate an entity, such as associating the \textit{flick} and \textit{film} keyword with the \textit{Movie} entity. It is our belief that this corpus will lead to improved results as it allows us to better adjust our system's expectations.   


\section{Tools and Methodology}
\label{sec:methodology}
In this section, we describe the tools and methodology we use to build SlugNERDS.

\subsection{Google Knowledge Graph}
\label{sec:gkg}

\begin{figure}[h!]
  \centering
    \includegraphics[width=\linewidth]{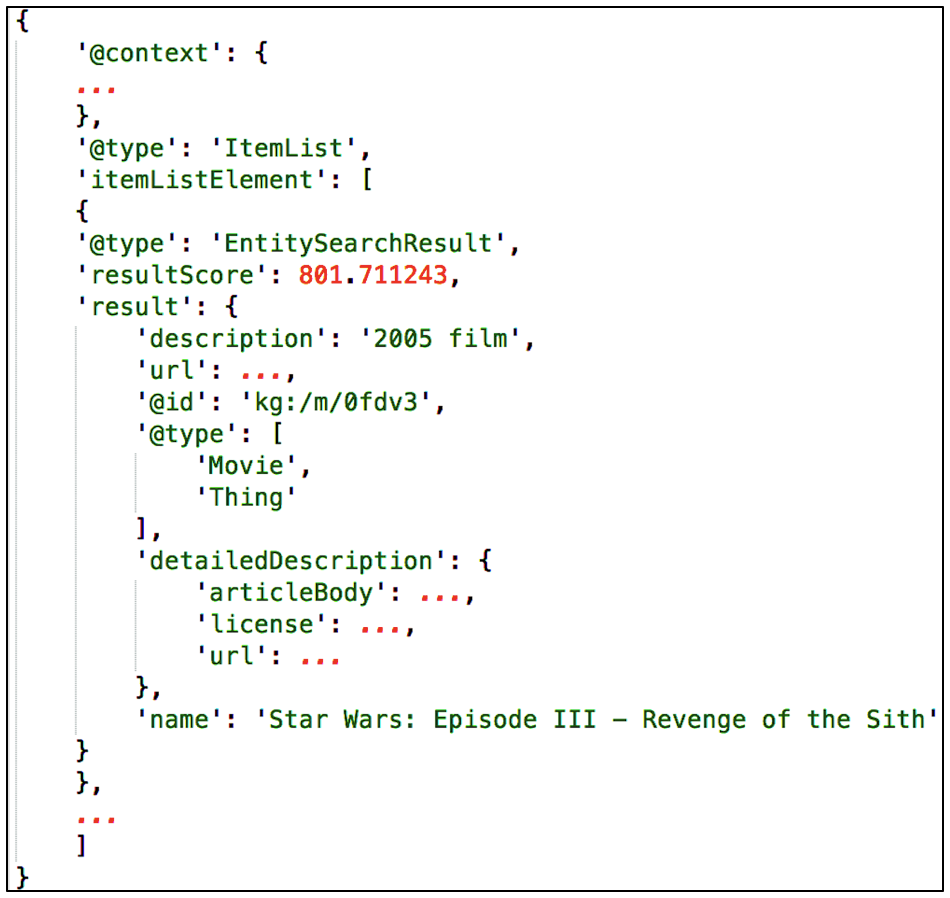}
    \caption{Google Knowledge Graph Search API Result For the Query \textit{revenge of the sith}}\label{fig:gkg}
\end{figure}

The SlugNERDS tool primarily utilizes the Google Knowledge Graph API \footnote{\url{https://www.google.com/intl/bn/insidesearch/features/search/knowledge.html}} to identify known entities. A query to the Google Knowledge Graph API returns a list of the $N$ most probable entities which are sorted based on an associated base score. This score is provided by the API and is assumed to be a combination of contextual overlap and entity popularity. Each entity has substantial meta data including the full name of the entity, Schema.org type classification, and both a brief and long description with an associated Wikipedia article. Having automatic access to an associated Wikipedia article allows us to reliably solve the task of web based NEL contingent on successfully performing NER. An example query result can be seen in Figure \ref{fig:gkg}.

Utilizing this API to identify known entities is ideal as it isn't domain specific, and requires no training from users of the tool. While directly querying the API in real time may sound expensive, this tool has been successfully deployed in our real time conversational agent, Slugbot, without significantly inhibiting the user experience. 

\subsection{Schema.org}
\label{sec:schema}

The Google Knowledge Graph  API classifies entities using the Schema.org \footnote{\url{https://www.schema.org}} entity ontology. Schema.org is an effort to create a richer web infrastructure by proposing common MicroData for entities within a website. Millions of websites which contain rich structured data across an array of subjects - such as IMDb \footnote{\url{https://www.imdb.com}}, BestBuy \footnote{\url{https://www.bestbuy.com}}, BarnesNNobles \footnote{\url{https://www.barnesandnoble.com}}, and Yelp \footnote{\url{https://www.yelp.com}} - have already adopted this MicroData in some form.

This not only enforces consistency across a multitude of data streams, but allows us to connect common entities to their related attribute types
(such as {\it SportsTeam} $\rightarrow$ {\it athlete} $\rightarrow$ {\it Person} $\rightarrow$ {\it birthDate}), allowing the system to retrieve a large set of possible
next topics, related facts, and associated entities. We can further expand on potential topics by utilizing the schema ontology to access properties of the entities higher up in the hierarchy.

\subsection{SlugNERDS Pipeline}
\label{sec:ourMethod}
\begin{figure}
  \centering
    \includegraphics[width=\linewidth]{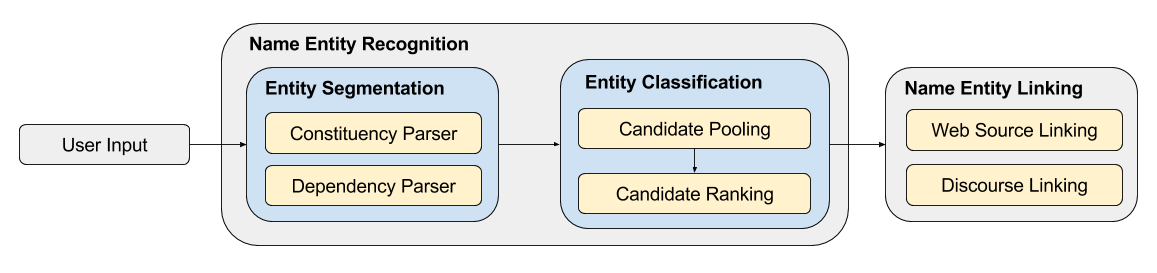}
    \caption{The SlugNERDS pipeline.}\label{fig:pipeline}
\end{figure}

Figure \ref{fig:pipeline} represents the general SlugNERDS pipeline. Our \emph{Name Entity Recognition} consists of two standard modules, \emph{Entity Segmentation} and \emph{Entity Classification} \cite{Ritter11TwitterNER,Collins99unsupervisedmodels,DowneyBE07entityinweb}. Subsequently we perform \emph{Entity Linking} on the recognized entity. We will examine this process with the following example utterance: \textit{"I think my favorite star wars is revenge of the sith"}. Please note that punctuation is not included in the user utterance.

\subsubsection{Entity Segmentation}\label{sec:chunk}

In order to refine our list of candidate strings to query we must break our text into reasonable chunks. We utilize a two pass approach to maximize our recall with a decent number of candidates. We maximize recall because a properly structured dialogue system will be able to pick the contextually relevant entity for follow-up questions and ignore extraneous entities which may have been misclassified. Furthermore, through empirical evaluation we have concluded that Slugbot missing an entity can be more detrimental to a conversation than over-classifying entities.

\begin{figure}[ht]
 \includegraphics[width=0.85\linewidth]{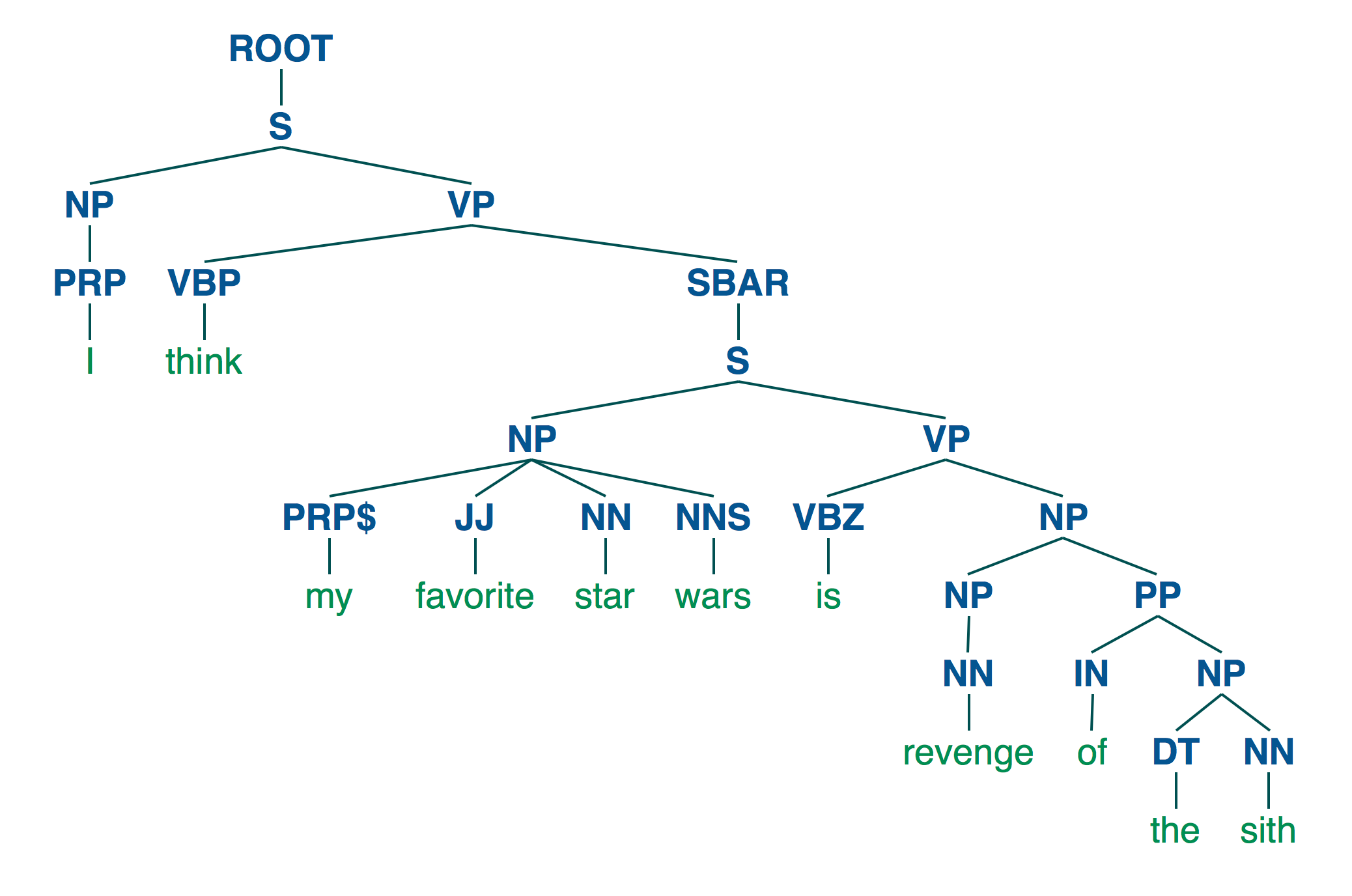}
 \caption{Constituency Tree}
 \label{fig:constituency_tree}
\end{figure}

First, we construct a constituency tree using Stanford CoreNLP \cite{corenlp} and build our candidate pool by collapsing each of the noun phrases, verb phrases, and sentences in the tree. A sample constituency tree can be seen in Figure \ref{fig:constituency_tree}. Additionally, we collapse sequential noun clusters from the dependency parse which have not yet been associated with an entity to create a secondary pool of candidates, so as to include more candidate strings that are ignored by shallow parsing \cite{Ritter11TwitterNER}. We exclude single pronouns such as \textit{I} and \textit{me} unless they seem extremely contextually relevant, such as in the case of \textit{It}\footnote{\textit{It} is currently a very popular horror movie.}, when discussing movies with the user. 

\subsubsection{Entity Classification}\label{sec:resolution}

Once we collect the candidate phrases in the Segmentation phase, we then query each of these phrases using the Google Knowledge Graph API and collect the top 6 relevant entities in \emph{Candidate Pooling}. Here 6 is an empirically derived value which represents a good range of possible candidates without including excessive candidates. Sometimes it is possible for the entity candidate returned by the query to have the same exact title with different entity types. For example, the phrase \textit{revenge of the sith} returns 5 candidates with the title \textit{Star Wars: Episode III - Revenge Of The Sith} each with a different entity type (\textit{Movie}, \textit{Video Game}, \textit{Book}, \textit{MusicAlbum}, and \textit{BookSeries}). The base score can vary significantly between these versions, for example the base score for the \textit{Movie} entity is 795.59, while it is 138.05 for the \textit{BookSeries} entity - this is assumed to be due to the popularity of the movie vs. the book series.

\begin{table}[ht]
\begin{small}
\centering
\begin{tabular}
{|p{.45\textwidth}|}
\toprule
{\bf User Utterance:} I think my favorite \emph{Star Wars} movie is \emph{revenge of the sith}.\\
{\bf Candidate:} Star Wars: Episode III - Revenge of the Sith $['Movie', 'Thing']$\\
{\bf Candidate Initial Vector:} $[1,1,1,1,1,1,1,1]$\\
{\bf Candidate Overlap:} $[1,1,0,0,0.11,.75,.75,1]$\\
\bottomrule     
 \end{tabular}
 \caption{Example word vector translation.}\label{table:slug_ner}
\end{small}
\end{table}

Since the entities returned by the Google Knowledge Graph may not be an exact match to our query, it gives us more flexibility, while introducing some noise. Furthermore, if we are expecting a user to talk about certain entity types according to the context, as discussed when introducing the SchemaActuators corpus in Section \ref{sec:overview}, we place increased value on certain entities while penalizing others. Thus, we perform our scoring algorithm to maximize our performance in the \emph{Candidate Ranking} phase. 

First, we create a word overlap vector indicating the total overlap. Because we expect noise in the user's utterance, we allow for unexpected words to be inserted in the middle of the phrase realizing the entity with a distance based penalty. 
Table~\ref{table:slug_ner} gives an example of the word vector. The word vector is initialized as 1 for each candidate token, for instance, \emph{Star Wars: Episode III - Revenge of the Sith} has 8 tokens, thus the initial vector is [1,1,1,1,1,1,1,1]. Then we compare the candidate tokens with the user utterance. The element of the vector will be penalized if tokens are missing or extra tokens are detected. For example, the element 3 and element 4 are set to zero because there is two missing words (\textit{Episode III}), resulting in the 5th element of the vector receiving a penalty of $\frac{1}{3}$, which is the distance between the matching tokens. There are also two extra words (\textit{movie is}) in the user utterance, thus an additional penalty of $\frac{1}{3}$ is applied to the 5th element, which yields a 0.11 score. We also note the imposed penalty on stop words by multiplying the element with 0.75 - this helps to increase our precision as we try to prevent correctly positioned stop words from accidentally forming an entity. The 6th and 7th elements (\textit{of the}) are the stop word examples. Equation \ref{eq:overlap} demonstrates how we use this word vector to alter the score of an entity.\\ 
\begin{equation}\label{eq:overlap}
\begin{aligned}
overlap\_score = sum (word\_vector) * \\
(\frac{1}{word\_vector.count (0) + 1}) * base score
\end{aligned}
\end{equation}

Secondly, we account for any entities we are expecting based on early inspection of the utterance's context:\\ 
\begin{equation} \label{eq:exp}
\begin{aligned}
can_score = overlap\_score * (num_matches + 1)
\end{aligned}
\end{equation}

Once the entities are all scored, we rerank our list and consider only the top ranked entity for each node, while also pruning away nodes whose top scoring entity was less than a certain threshold (empirically driven). Finally, we merge overlapping nodes who have candidates. For example, if both \textit{revenge of the sith} and a child node \textit{revenge} have the same entity as their top scoring entity, we will merge these two nodes or remove lower ranking conflicts.

In our last stage, we sync the query/candidate to our internal discourse state representation. In our example, two entities are extracted, mapping \textit{star wars} to entity type \textit{MovieSeries}, and \textit{ revenge of the sith} to \textit{Star Wars: Episode III - Revenge Of The Sith} with entity type \textit{Movie}.





\subsubsection{Entity Linking}\label{sec:link}
\emph{Named Entity Linking} is primarily encapsulated in two phases, \emph{Web Source Linking} and \emph{Discourse Linking}. With Web Source Linking we are interested in linking a known entity to existing resources on the web while discourse linking is focused on linking  each mention of the entity within the input to the same discourse  entity in our internal representation \cite{BFP87,WPJ97}.


As mentioned in Section \ref{sec:gkg}, the Google Knowledge Graph query returns a Wikipedia article associated with the entity. We can further increase our web based linking by utilizing the fact that a large number of popular websites use the Schema.org MicroData, allowing us to easily target relevant sources for information extraction. Finally, through empirical examination, we note that pairing the entity type with the precise entity name as provided in the query will allow for easy subsequent queries to large databases such as YAGO \cite{Rebele2016} or DBpedia \cite{DBPedia}.

Finally, our tool uses an augmented version of the Stanford Coreference Annotator \cite{corenlp} to perform Discourse Linking.

\section{Evaluation}
\label{sec:eval}

\begin{table*}[h]
\centering
\begin{tabular}{l|l|l|l|l}
\toprule
                 & Accuracy & macro-F1 & micro-F1 & weighted-F1  \\ \hline
Stanford Baseline & .029 & .079 & .029 & .052\\ \hline
Text Segmentation Only & .751 & .785 & .722 & .831 \\ \hline
SlugNERDS Scoring & .777 & .746 & .749 & .850 \\ \hline
SlugNERDS Scoring + Act\_Verbs & .770 & .752 & .747 & .849\\ \hline
SlugNERDS Scoring + Act\_Nouns & .773 & .762& .745 & .848 \\ \hline
Ensemble & .763 & .747 & .742 & .845\\
\bottomrule         
\end{tabular}
\caption{NER Results for the Threshold 0 experiment.}
\label{table:ner-results}
\end{table*}

\subsection{NER Results}
\begin{table*}[h]
\centering
\begin{tabular}{l|l|l|l|l}
\toprule
                 & Accuracy & macro-F1 & micro-F1 & weighted-F1  \\ \hline
Stanford Baseline & .029 & .079 & .029 & .052 \\ \hline
Text Segmentation Only & .521 & .656 & .511 &  .663\\ \hline
SlugNERDS Scoring & .553 & .624 & .546 & .692 \\ \hline
SlugNERDS Scoring + Act\_Verbs & .604 & .655 & .598 & .735\\ \hline
SlugNERDS Scoring + Act\_Nouns & .579 & .619 & .571 & .714\\ \hline
Ensemble & .592 & .634 & .586 & .726 \\
\bottomrule         
\end{tabular}
\caption{NER Results for the Threshold 150 experiment.}
\label{table:150-ner-results}
\end{table*}



Our SlugNERDS tool was originally developed and utilized in the 2017 Alexa Prize Competition for SlugBot, which scored in the top 25\% of competing social bots. For detailed system evaluation, we present a set of experiments to evaluate our SlugNERDS tool and SchemaActuator corpus, independently from SlugBot. We evaluate SlugNERDS using (1) the base scores using only text segmentation with the Google Knowledge Graph, (2) augmenting the scores using our scoring algorithm, (3) using the SchemaActuator actions to increase context, (4) using the SchemaActuator keywords to increase context, (5) an ensemble approach. Moreover we test these configurations with two different scoring thresholds (the minimum score required for an entity be accepted as the correct class). The two thresholds we will test are (1) 150, an empirically driven value deemed to be a reasonable threshold during the development of SlugBot and (2) 0, no threshold at all.


Table \ref{table:ner-results} presents the results of our NER experiments using the SchemaEntityDB. Since Stanford NER is still commonly used in many state of the art open domain conversational systems, we use it as our baseline system. Our other experiments include using just our text segment method, then adding SlugNERDS ranking, iteratively adding in verbs and nouns from our SchemaActuators corpus, and finally showing our ensemble method results (which merges all resources). Since Stanford provides coarse grain entity type such as PERSON, LOCATION, ORGANIZATION, and MISC, we map the specific gold standard label to these four types for the evaluation. 

We use accuracy, macro-f1, micro-f1, and weighted-f1, to have a better understanding of our system's performance. High accuracy entity detection will ensure a more satisfiable conversation, and prevent us from missing the topics of the conversation. The macro-f1 treats all the classes evenly, the micro-f1 accounts for label imbalance, and the weighted-f1 is a weighted macro-f1. Since detecting in-frequent entities is also important for us, we evaluate the macro-f1, the micro-f1 and the weighted-f1 for different interests. Table \ref{table:ner-results} shows that Stanford yields a macro-F1 score as 0.079. As predicted, we see from these results that Stanford NER is not a suitable system to use when detecting entities in open domain discourse. Our text segmentation model has an accuracy of 0.751, macro-F1 of 0.785, and weighted-F1 of 0.831. Our SlugNERDS models has a better accuracy 0.777 and a slightly worse macro-F1 of 0.746, but a better weighted-F1 0.85. After utilizing the SchemaActuator corpus, the macro-F1 are improved slightly, though the weighted-F1 is almost the same, which might due to the reason that the extra Act\_Verbs and Act\_Nouns are able to detect the edge cases which are in-frequent entities. We believe that when integrated with a real dialogue system, the increase in contextual knowledge from our dialogue manager will yield further increase performance.

In our original experiment we see that a low threshold doesn't need contextual information to classify entities. However after examining the results it's clear that while more correct entities are being classified, there is also increased levels of over-classification which can be detrimental to the system. Therefore we aim to increase our accuracy with an with an empirically derived threshold of 150. In this experiment our model continued to outperformed the stanford baseline and the text segmentation results were the worst configuration for SlugNERDS. We can also see by comparing Table \ref{table:ner-results} and Table \ref{table:150-ner-results} that as we increase this threshold, we increase the impact of adding contextual information to SlugNERDS. This implies the importance of encoding contextually relevant data in our model while also increasing our tolerance of noise. In future work we aim at further analyzing our distribution of annotated entities to pick an optimized threshold value. 


\subsection{Distribution of Detected Entities}
\begin{figure}[h!]
  \centering
    \includegraphics[width=\linewidth]{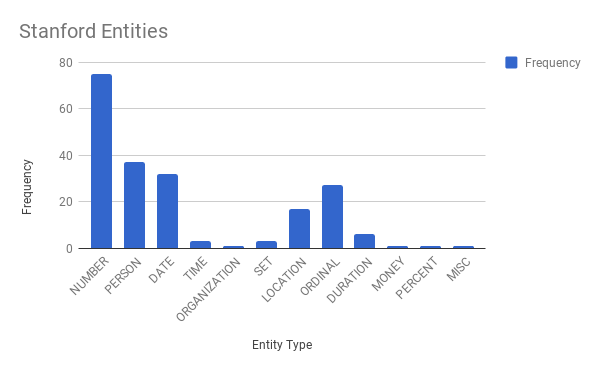}
    \caption{Distribution entity types detected by the Stanford NER.}\label{fig:stanford_entity_distribution}
\end{figure}

\begin{figure}[h!]
  \centering
    \includegraphics[width=\linewidth]{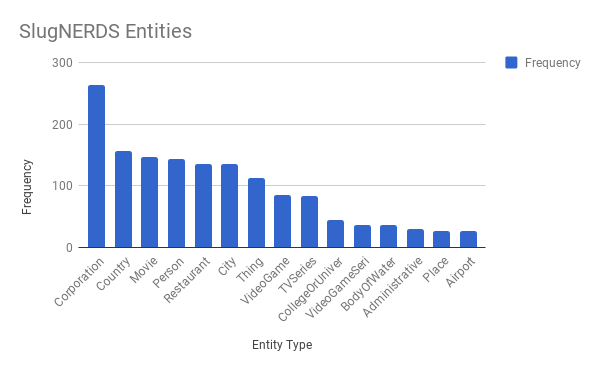}
    \caption{Distribution of top 15 entity types that our model detected.}\label{fig:slug_entity_distribution}
\end{figure}

\begin{figure}[h!]
  \centering
    \includegraphics[width=\linewidth]{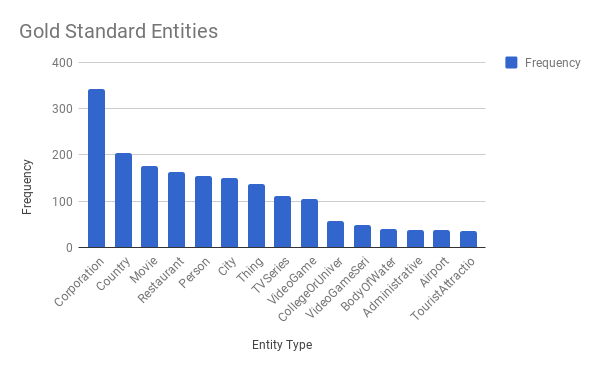}
    \caption{Distribution of top 15 entity types in the annotated data.}\label{fig:gold_entity_distribution}
\end{figure}

Figure~\ref{fig:stanford_entity_distribution} shows the distribution of Stanford Entities that are detected. The NUMBER, DATE, TIME, SET, ORDINAL, DURATION, MONEY, and PERCENT are not very useful to our system as we are interested primarily in known entities. Therefore we are most interested in the PERSON, ORGANIZATION, LOCATION, and MISC types - however, all of these entity types detected much less frequently than they appear in our annotated data, as can be seen in Figure \ref{fig:gold_entity_distribution}. Figure~\ref{fig:slug_entity_distribution} shows the top 15 most frequent entities that caught our SlugNERDS Models. We can see that our SlugNERDS model is able to successfully detect large amounts of various conversational entities such as Movie, Book, and MusicRecording, while also maintaining a very similar distribution as seen in our annotated data.



\subsection{Error Analysis}
Here we will discuss different errors we noticed while analyzing the results of the experiments described above. Primarily, we see three different classes of errors; insufficient contextual information, natural language understanding deficiencies, and the difficulty of encapsulating every possible entity within the Google Knowledge Graph.

Table \ref{table:context_errors} demonstrates several errors which are the result of insufficient contextual information. While this evaluation was meant to analyze SlugNERDS as a stand-alone tool, it is clear that these errors can be resolved by enhancing the contextual knowledge we pass to our tool. For example, while there are no lexemes which indicate \textit{lord of the rings} in Sample \ref{subtable:entity_overlap} is referring to the MovieSeries, the state of our dialogue system could inform us that we are discussing movies, rather than books. As in Sample \ref{subtable:series_expansion}, we see that it common in colloquial speech to refer to an element of a series by it's common root word. For example \textit{halo} in the expression \textit{let's play halo} is valid for \textit{halo combat evolved}, \textit{halo 2}, \textit{halo wars}, and various other titles in the series. Disambiguating this is not a trivial problem and requires a significantly more rich state than the previous example. We will note here that it is possible to also leverage the granularity of our Schema.org entity types to recognize we are talking about a VideoGameSeries and clarify the specific VideoGame with the user. Finally, Sample \ref{subtable:user_context} demonstrates how a lack of user meta-data, in this case location and frequency, results in incorrect classifications. A dialogue system can represent this contextual data by referencing a user model which may indicate that since the user lives in California, they are likely referring to San Jose, California, rather than San Jose, Costa Rica.

Table \ref{table:nlu_errors} demonstrates two areas in which adding an additional layer of NLU would yield increased performance. Specifically we notice in Sample \ref{subtable:abbreviations} that abbreviations cause difficulty in classifying entities. While ucsd will in-fact return the correct entity as a potentially candidate from the Google Knowledge Graph, there is no lexical overlap, resulting in a false classification. By adding an additional layer of NLU which is able to expand abbreviated entities we would see an increase in performance. As described previously, the SlugEntityDB annotated corpora was designed assuming the input provided is from a spoken dialogue system - meaning it will suffer from the limitations of state of the art automatic speech recognition. More specifically, our utterances have no punctuation, capitalization, or non-alpha-numeric symbols. While we have already resolved this partially such that utterances like \textit{x man} will correctly map to \textit{X-man}, the lack of punctuation and capitalization can lead to inaccurate results from our parser - this can directly alter the queries which are sent to the Google Knowledge Graph. This can be seen in Sample \ref{subtable:parser_fail}, where the best possible query for the original utterance was \textit{love gordon ramsey} which does not yield the correct entity. After manually capitalizing the entity name, our best possibly query became \textit{Gordon Ramsey}, which resulted in the correct entity. 

Finally, as there exists an infinite amount of entities, it is reasonable to surmise that the Google Knowledge Graph, while quite robust, is incomplete. For example, querying either \textit{xbox 360} or \textit{windows} both return the \textit{Microsoft} entity has the best possible candidate. While not necessarily common, this can be a difficult problem to deal with, especially when an entire category of entities, such as operating systems or video game consoles, are missing from the Knowledge Graph. 

\begin{table}
\begin{small}
\begin{subtable}[h!]{1.0\linewidth}
\begin{tabular}
{@{}p{3.4cm}|p{2.2cm}|p{2.2cm}@{}}
\toprule
Utterance & Correct Entity & Predicted Entity \\ \hline
the lord of the rings was my childhood & MovieSeries & BookSeries \\
\bottomrule        
 \end{tabular}
\caption{Without sufficient contextual knowledge, it is difficult to differentiate between entities with the exact same title.}\label{subtable:entity_overlap}

\vspace{.3cm}

\begin{tabular}
{@{}p{3.4cm}|p{2.2cm}|p{2.2cm}@{}}
\toprule
Utterance & Correct Entity & Predicted Entity \\ \hline
halo has been dead for a while now & VideoGame, Halo: Combat Evolved & VideoGameSeries, Halo)\\ 
\bottomrule        
 \end{tabular}
\caption{Entities which are nested within a series can also be hard to detect, such as here when the user is talking about the first Halo game vs. the Halo series.}\label{subtable:series_expansion}

\vspace{.3cm}

\begin{tabular}
{@{}p{3.4cm}|p{2.2cm}|p{2.2cm}@{}}
\toprule
Utterance & Correct Entity & Predicted Entity \\ \hline
sacramento airport was pretty busy& Sacramento International Airport & Sacramento Airport\\ \hline
i did not know san jose is a capital& San Jose, California & San Jose\\ 
\bottomrule        
 \end{tabular}
\caption{Without contextual information about the user, it's difficult to differentiate between San Jose refers to San Jose California vs. San Jose Costa Rica. Similarly, "common sense" indicates that "Sacramento Airport" refers to Sacramento International Airport, rather than the much smaller Sacramento Airport.}\label{subtable:user_context}

\end{subtable}
\end{small}
 \caption{Common errors stemming from insufficient contextual knowledge.}
 \label{table:context_errors}
\end{table}

\begin{table}
\begin{small}
\begin{subtable}[h!]{1.0\linewidth}
\begin{tabular}
{@{}p{3.4cm}|p{2.2cm}|p{2.2cm}@{}}
\toprule
Utterance & Correct Entity & Predicted Entity \\ \hline
i love gordon ramsay& Person & None\\ \hline
i love Gordon Ramsay& Person & Person\\
\bottomrule        
 \end{tabular}
\caption{A lack of capitalization and punctuation leads to parser errors, resulting in missed queries.}\label{subtable:parser_fail}

\vspace{.3cm}

\begin{tabular}
{@{}p{3.4cm}|p{2.2cm}|p{2.2cm}@{}}
\toprule
Utterance & Correct Entity & Predicted Entity \\ \hline
ucsd is number 23& University of California, San Diego & None \\
\bottomrule        
 \end{tabular}
\caption{Mismatch between the string in the utterance vs the actual entity name, as in the case with an abbreviation, causes classification to fail. }\label{subtable:abbreviations}

\end{subtable}
\end{small}
 \caption{Common errors stemming from gaps in our natural language understanding pipeline.}
 \label{table:nlu_errors}
\end{table}

\section{Conclusion and Future Work}
\label{sec:conc}
In this paper we have presented SlugNERDS, an NER and NEL tool which is optimized with the respect to the challenges that are specific to open domain conversation. We have also presented two relevant corpora, one being the first dataset of its kind to be annotated with Schema.org named entity types in addition to the SchemaActuator corpus - a mapping of actions and keywords to their respective Schema.org entity types.

To our knowledge our system is the only one to utilize the Schema.org entity types for entity classification, but we plan to compare our system more extensively to other existing NER systems in future work, which may allow us to improve our system. One is T-NER System \cite{Ritter11TwitterNER} \footnote{\url{https://github.com/aritter/twitter_nlp}}, which is optimized for NER in Tweets, as discussed in Section \ref{sec:intro}. We are also interested in comparing against a state of the art neural NEL model \cite{GuptaSiRo17RothNEL} \footnote{\url{https://nitishgupta.github.io/neural-el/}}. This model more proactively tackles the open domain problem, and can include updated knowledge of existing entities without retraining. There are clearly many areas in which we can further evaluate and improve the performance of SlugNERDS. Specifically, we plan to evaluate the performance of SlugNERDS with an emphasis on the impact of contextual information provided as a result of our dialogue system's state tracking capabilities. We also plan to improve our natural language understanding pipeline by investigating methodologies for improving input text quality, such as automatic capitalization and punctuation insertion. Finally, we are currently investigating how reinforcement learning can be applied to our existed conversational data to increase our models contextual insight.

\section{References}
\label{sec:ref}

\bibliographystyle{lrec}
\bibliography{xample,nl}

\end{document}